\documentclass{article}

\usepackage[nonatbib,final]{neurips_2020}
\usepackage{graphicx}
\usepackage{subfigure}
\usepackage[utf8]{inputenc} 
\usepackage[T1]{fontenc}    
\usepackage{hyperref}       
\usepackage{url}            
\usepackage{booktabs}       
\usepackage{amsfonts}       
\usepackage{nicefrac}       
\usepackage{microtype}      
\usepackage{cite}
\usepackage{hyperref}

\title{Detection of Malaria Vector Breeding Habitats using Topographic Models}

\author{%
  Aishwarya Jadhav\\
  Omdena\\
  \texttt{me.aishwaryajadhav@gmail.com} \\
}

\begin{document}

\maketitle
\begin{abstract}
Treatment of stagnant water bodies that act as a breeding site for malarial vectors is a fundamental step in most malaria elimination campaigns. However, identification of such water bodies over large areas is expensive, labour- intensive and time-consuming and hence, challenging in countries with limited resources. Practical models that can efficiently locate water bodies can target the limited resources by greatly reducing the area that needs to be scanned by the field workers. To this end, we propose a practical topographic model based on easily available, global, high-resolution DEM data to predict locations of potential vector-breeding water sites. We surveyed the Obuasi region of Ghana to assess the impact of various topographic features on different types of water bodies and uncover the features that significantly influence the formation of aquatic habitats. We further evaluate the effectiveness of multiple models. Our best model significantly outperforms earlier attempts that employ topographic variables for detection of small water sites, even the ones that utilize additional satellite imagery data and demonstrates robustness across different settings.
\end{abstract}

\section{Background}
An endemic in 31 countries, malaria claimed over 405,000 lives with global case counts exceeding 228 million in the year 2018. Of these, 93 per cent of the cases were encountered in Africa \cite{who}. Recent analysis \cite{fillinger} suggests that new interventions need to be added to the front-line vector control tools that primarily target adult mosquitoes. According to this study, Larval Source Management (LSM) is a highly effective method of malaria control and can reduce transmission by $70-90\%$ in settings where mosquito larval habitats are well-defined. However, the major challenge in the adoption of LSM in Africa is the high number of small and temporary larval habitats making them so extensive that not all of them can be covered on foot. Hence, there is a need for defining the targets using models that can efficiently detect such water bodies while utilizing minimum resources so as to not strain the already resource-extensive task of larviciding. 

Many previous studies have underscored the importance of topographic features for modelling the spatial distribution of malarial risk\cite{atieli,cohen1,clennon}. Some attempts have studied the distribution patterns between clinically diagnosed cases of malaria, household features and their surrounding topographies\cite{haque}. However, these are not scalable as moving to a new region with no prior clinical or household survey data at hand is not an easy task. Other studies have explored the influence of landscape features on the formation of aquatic habitats that serve as vector-breeding sites. Some of these studies employ different satellite imagery data (LANDSAT, IKONOS) to derive features such as land-cover, land-use and vegetation indices to use in conjunction with topographic features \cite{clennon,munga,mccann}. Clennon et. al. have achieved a ROC of 0.81 for the prediction of water sites using LANDSAT and DEM data \cite{clennon}. However, for low-resource regions, the availability of such imagery with an acceptable quality can be expensive or not possible. Until 2014, even DEM data of resolution better than 90m was not available for the African continent. Besides, the generation of complex features and indices is time-consuming and heavy-weight for practical purposes over large regions. Nmor et. al. have used just topographic variables in their Multiple Logistic Regression (MLR) model for the prediction of habitats and have been able to achieve a ROC of 0.83 with SRTM 90m resolution DEM data \cite{nmor}. 

In this study, we developed models for detecting the locations of stagnant water bodies using only the topographic features derived from a higher resolution (30m) version of the SRTM DEM data. We assessed the effectiveness of the topographic features highlighted by multiple earlier studies, identified newer features which show significant correlation with water source locations and delineated features, indicated by previous studies, that no longer weigh in on the prediction results. We have used the survey results of a 140 km\textsuperscript{2} region to conduct this study, which is much larger than any of the previous attempts (<60 km\textsuperscript{2}); hence we expect our model to be viable across varied landscapes.
\section{Methods}
\subsection{Survey Design}

\paragraph{Study Site} Our study region, bounded by -1.74 to -1.57 longitude and 6.14 to 6.25 latitude, spans an area of 230 km\textsuperscript{2} and covers the town of Obuasi in the Ashanti Region of Ghana. Obuasi has semi-equatorial climate and two rainy seasons extending from March to July and from mid-September to December. It has a mean annual rainfall between 125mm and 175mm and average yearly precipitation of over 1450 mm. Obuasi is an urban community with over 175,000 people.

\paragraph{Site Survey} A ground survey of around 60 per cent of the bounded region, that covered an area of $\sim$140 km\textsuperscript{2}, was conducted at the end of the wet season between July and August. The region was divided into MGRS chunks of size 100mx100m. Field crew surveyed all the planned chunks for water sites. GPS was used to track the positions of the field crew. Chunks that were scanned more than 70 per cent but where no water body was found were marked as negative. Whenever a water body was found, the field workers photographed it and categorized it into one of the following categories: tracks, swamp, puddle, pool, pond, fringe, footprint, construction, drainage canal, other. Also, the location of this water source was recorded via GPS.

\subsection{Survey Outcomes}

\paragraph{Positive Sources}
The survey found 6214 water bodies. Of these, 2988 were classified as natural water formations (swamp, puddle, pool, pond, fringe). These were included as positive samples in dataset A. Previous studies have also included water sources such as canals, footprints and tracks into their analysis reasoning that the formations or construction of these may be correlated to topographic features of the land \cite{nmor}. Thus, we created dataset B consisting of natural water bodies plus canals, tracks and footprints as the 4216 positive water sources.

\paragraph{Negative Sampling}
The survey resulted in 944 negative MGRS chunks. To generate negative samples, we picked points from the negative chunks that adhered to the following criteria: each negative point must be at a distance of at least 100m from any positive point and the distance between any two negative points must be at least 30m. The distance constraint between the positive and negative points was imposed to allow sufficient variations in the topographic features of the two opposing locations enabling the model to learn appropriately. Negative points were required to be 30m apart to avoid them from being in the same cell of the DEM grid of resolution 30m that would be used to create the topographic features. To achieve this, we overlay the negative MGRS chunks on the DEM grid and extracted the centres of the cells that lay in the negative chunks. We could generate $\sim12000$ such negative points, out of which, 6566 points satisfied the distance constraint with the positive points. These negative samples were then added to datasets A and B.

\section{Topographic Features}
\paragraph{DEM Data}
We calculated all the topographic features using the Shuttle Radar Topography Mission (SRTM) version 3, void-filled Digital Elevation Model (DEM) data. SRTM elevation imagery was collected during a space shuttle mission in 2000 using a multi-frequency and multi-polarization radar system \cite{srtm}. This dataset is present in raster format, where each pixel represents elevation data over a resolution of 1 arc-second or 30mx30m. This dataset was made available globally, at no cost, in 2015. Prior to that only the 90m resolution data was publicly available for regions outside the USA.

\paragraph {Topographic Variables}
Eighteen topographic variables were created from the DEM data using the open-source SAGA API \cite{saga}. Before the creation of these features, the DEM elevations were pre-processed to fill unexcepted sinks or spikes that could potentially represent errors, using the method described by Wang et. al. \cite{wang}. We then conducted basic terrain analysis to generate the features.

\paragraph {Feature Analysis}
In order to determine the features that have the most significant impact on the presence or absence of water bodies, we performed univariate logistic regression analysis on them to generate respective P-values. For dataset A, we found the following features to be the most indicative: Topographic Position Index (TPI500), Topographic Wetness Index (TWI), Channel Network Distance (CND), Relative Slope Position (RPS), Closed Depressions and Flow Direction.
 
TPI at a point is the difference between the elevation of the point and the mean elevation of neighbouring points \cite{weiss}. We considered a neighbourhood radius of 500m (TPI500), which captures local hills and depressions. TWI is a measure of the moisture retention capability of the soil. It is calculated as the ratio of the upslope contributing area to the tangent of the local slope. CND captures the distance between a point and the closest natural channel. RPS is a combined terrain parameter calculated using Altitude Above Channel Lines (AACL) and Altitude Below Ridge Lines (ABRL) as the ratio of AACL to (AACL+ABRL)\cite{bock}. Closed Depressions are depression areas that are covered with lakes or bogs. Finally, flow direction is the direction of water flow through each cell in the DEM grid.

For dataset B, we found that along with TPI500, TWI, CND and RPS, three other features showed significant weightage: Plan Curvature, Profile Curvature and Convergence Index (CI). Plan Curvature is the curvature in a horizontal plane. It is positive for concave contours and negative for convex. Profile Curvature is the curvature of the surface in the direction of the steepest slope in the vertical plane of a flow line. CI shows the structure of the relief as a set of convergent areas (channels) (CI=1 to 100) and divergent areas (ridges) (CI=-1 to -100).

As in most of the previous studies, TWI and TPI were found to be prominent features for determining the presence of water bodies. Moreover, as indicated by Nmor et.al., plan and profile curvatures do affect the locations of water bodies, but only in dataset B; they have little significance for natural water sites in dataset A. Furthermore, although the earlier studies indicated slope as an important factor, our analysis found slope to be moderately correlated to the water-body locations in both datasets. Interestingly, complex features such as RPS and TWI that are derived using slope show a higher correlation to the target variable. Finally, aspect, that was indicated by previous studies as a crucial feature showed very little correlation with the target variable in both datasets. 

To determine mutual correlations between the significant features, we used Pearson’s correlation coefficient. Features that have high mutual correlations (absolute value>0.85) were compared based on their Mutual Information (MI) values with the target variable. We found a high correlation between RPS and CND. The later was not used in the classification models based on the MI scores of the two.

\section{Models}

To classify a cell of a DEM grid as positive or negative for water bodies, we trained and evaluated different models using both the datasets. Logistic Regression, Linear SVM, Extra Trees Classifier, Random Forest Classifier and Gradient Boosting Classifier models were evaluated. We engaged a variety of model types to assess linear as well as non-linear relationships between the features and the target variable. The models were evaluated using the metrics: AUC of Receiver Operating Characteristics (ROC), Recall/Sensitivity, Specificity and Precision. Models were optimized to have a high recall so as to miss as few positive cells/water bodies as possible. The datasets were split into train and test sets based on the 80th percentile of the longitude of the points as threshold. All the points on the west of this threshold were incorporated in the train set; those on the east were used for validation. Due to the higher number of negative points in the train and test sets, class weights were assigned as ‘balanced’ during training and negative points were under-sampled while testing.

\section{Results}

\paragraph{Dataset A}The evaluation of the models on 600 positive and 600 negative samples yielded a ROC > 0.87 for all the models. The best performing model, Gradient Boosting Classifier (GBC), had a ROC of 0.90 with a recall/sensitivity of 0.94, precision of 0.70 and specificity of 0.60. The variables used were TPI500, TWI, RPS, Closed Depressions and Flow Direction. The classification suggests that small water sites are most likely to occur in the proximity of closed depressions (such as lakes) with high TWI and local minima of elevation (TPI) or flat slopes (RPS). The following table depicts the performances of the prior MLR model and all of our models.
\begin{table}[!h]
  \caption{Performance of the models on Dataset A}
  \label{datasetATable}
  \centering
  \begin{tabular}{llllll}
    \toprule
    \textbf{Model}     & {\textbf{ROC}}     & {\textbf{Recall/Sensitivity}}  & {\textbf{Precision}}   & {\textbf{Specificity}}  & {\textbf{F1-score}} \\
    \midrule
    {\textbf{MLR Model} \cite{nmor}}  &0.829   &0.815   &-   &0.739  &-  \\
    {\textbf{Random Forest Classifier}}  &0.88   &0.89   &0.74   &0.69  &0.81  \\
    {\textbf{Extra Trees Classifier}}      & 0.89   &0.90   &0.71   &0.63  &0.79 \\
    {\textbf{Gradient Boosting Classifier}}     &0.90   &0.94   &0.70   &0.60   &0.80 \\
    {\textbf{Linear SVM}}     &0.87   &0.91   &0.71   &0.63   &0.80 \\
    	
    {\textbf{Logistic Regression Classifier}}     & 0.87  & 0.90  &0.70   &0.61  &0.79 \\
    \bottomrule
  \end{tabular}
\end{table}

\paragraph{Dataset B} The test set consisted of 850 positive and negative sites each. Due to the presence of some artificial positive sources that are not strongly impacted by topographic variations, the performances of the models trained using this data were lower than those trained on dataset A. The best performing model, Random Forest Classifier (ROC=0.85), had a recall of 0.87 for a precision of 0.70 and a sensitivity of 0.63. The ROC of this model depicts a $3.7\%$ increase as compared to the baseline. The features used were TPI500, TWI, RPS, Plan Curvature, Profile Curvature and CI. 

\begin{table}[!h]
  \caption{Performance of the models on Dataset B}
  \label{datasetBTable}
  \centering
  \begin{tabular}{llllll}
    \toprule
    \textbf{Model}     & {\textbf{ROC}}     & {\textbf{Recall/Sensitivity}}  & {\textbf{Precision}}   & {\textbf{Specificity}}  & {\textbf{F1-score}} \\
    \midrule
    {\textbf{Random Forest Classifier}}  &0.85   &0.87   &0.70   &0.63  &0.78  \\
    {\textbf{Extra Trees Classifier}}      & 0.84   &0.85   &0.67   &0.58  &0.75 \\
    {\textbf{Gradient Boosting Classifier}}     &0.85   &0.86   &0.72   &0.67   &0.78 \\
    {\textbf{Linear SVM}}     &0.85   &0.86   &0.67   &0.58   &0.75 \\
    {\textbf{Logistic Regression Classifier}}     & 0.85  & 0.86  &0.67   &0.58  &0.75 \\
    \bottomrule
  \end{tabular}
\end{table}
\section{Discussion}
We present a simple yet effective tool for targeted LSM initiatives in low-resource settings that just uses freely available DEM data to generate risk maps of potential vector-breeding water sites. It does not require hard-to-collect clinical data or features that require complicated processing. As per our literature review, our best-performing model has been able to significantly outperform the existing models that detect small water bodies using topographic features. The use of higher resolution DEM data and identification of significant topographic features has played a major role in it. Apart from using the widely acknowledged features of TPI and TWI, we also used features such as RPS, Closed Depressions and Flow Direction which weren’t employed by any of the earlier studies. We believe that the higher resolution data helped uncover more nuanced patterns while the vastness of our survey region facilitated capturing a wider range of dependencies between the topographic variables and water accumulation. Our study region of $\sim$140 km\textsuperscript{2} is the largest so far; hence it consists of a much wider variety of topographies and breeding sites. Consequently, our models is likely to generalize well across newer regions with differing landscapes. Furthermore, contrary to the study areas of the baseline models, our survey region depicts an urban area. Urban regions with high populations and greater density of man-made structures have an impact on the surrounding natural ecologies and habitats therein; they add a certain degree of noise to the feature sets affecting the performance of the models. The robust performance of our model even under such settings provides a greater degree of confidence in its applicability across regions with varying population and topographic characteristics. 

\section*{Acknowledgements}
This study was undertaken as part of the Preventing Malaria Infections Through Topography and Satellite Image Analysis project sponsored by Omdena in collaboration with Zzapp. We would like to thank Zzapp for providing us with the survey data from Obuasi.

\bibliography{references}{}
\bibliographystyle{plain}
\end{document}